\documentclass{misc/nle}

\usepackage{amsmath}
\usepackage{graphicx}
\usepackage{natbib}
\ifpdf%
\usepackage{epstopdf}%
\else%
\fi
\usepackage[
    colorlinks=true,
    linkcolor=black,
    anchorcolor=black,
    citecolor=black,
    filecolor=black,
    menucolor=black,
    runcolor=black,
    urlcolor=black,
]{hyperref}
\usepackage{multirow}
\usepackage{times}
\usepackage{latexsym}
\usepackage[T1]{fontenc}
\usepackage[utf8]{inputenc}
\usepackage{microtype}
\usepackage{todonotes}
\usepackage{amsmath}
\usepackage{cleveref}
\usepackage{booktabs}
\usepackage{xcolor}
\usepackage{lipsum}
\usepackage{xspace}
\usepackage{enumitem}
\usepackage{mdframed} %
\usepackage{tcolorbox} %
\usepackage{ulem}
\usepackage{soul} %
\usepackage{float} %
\usepackage{wrapfig} %
\usepackage[labelfont={bf,sf,small},textfont=md,labelsep=space]{caption}
\setlist{noitemsep,topsep=0.5em}

\definecolor{TodoColor}{rgb}{1,0.7,0.6}
\definecolor{MartinColor}{rgb}{0.8,0.9,0.8}
\definecolor{VeraColor}{rgb}{0.4,0.9,0.8}

\newcommand{\ufalletter}{\hspace{-1mm}\text{
    \fontfamily{lmss}\selectfont
    \color{orange}{Ú}}
}
\definecolor{ethblue}{rgb}{0,0.1,0.4}
\newcommand{\ethletter}{\hspace{-0.5mm}\text{
    \fontfamily{phv}\fontseries{bx}\fontsize{7}{\baselineskip}\selectfont
    \textit{\textbf{\color{ethblue}{E}}}}
}

\newmdenv[
  linecolor=black,
  linewidth=1.2pt,
  topline=false,
  bottomline=false,
  rightline=false,
  leftmargin=0mm,
  innertopmargin=1mm,
  innerbottommargin=0.5mm,
  innerleftmargin=1mm,
  innerrightmargin=1.5mm,
  backgroundcolor=gray!10,
  skipabove=2.1\topsep,
  skipbelow=0.5\topsep,
]{quotebox}

\newcounter{textexamplecounter}
\crefname{textexamplecounter}{\protect{Example}}{\protect{examples}}
\Crefname{textexamplecounter}{\protect{Example}}{\protect{Examples}}

\newcommand{\textexample}[1]{
    \vspace{-5mm}
    \refstepcounter{textexamplecounter}
    \begin{quotebox}
        \textbf{Example: \thetextexamplecounter}
        
        \setlength{\parindent}{0cm}
        {
            \fontsize{8.7pt}{12pt}\selectfont
            #1
            
        }
    \end{quotebox}
}

\newcommand{\examplehighlight}[1]{%
\uline{#1}%
}

\def\pojem#1{\textit{#1}}

\newcommand{\hrefEmail}[2]{\href{mailto:#1}{\color{black}{#2}}}
\newcommand{\segmentseparator}{\textnormal{|||}\,}

\begin{document}
\label{firstpage}

\lefttitle{V. Zouhar, V. Kloudová, M. Popel, O. Bojar}
\righttitle{Evaluating Optimal Reference Translations}

\papertitle{Article}

\jnlPage{1}{00}
\jnlDoiYr{2024}
\doival{10.1017/xxxxx}

\title{Evaluating Optimal Reference Translations}

\begin{authgrp}
\author{Vilém Zouhar,$^{\ethletter\hspace{1mm}\ufalletter}$
Věra Kloudová,$^{\ufalletter}$
Martin Popel,$^{\ufalletter}$
and Ondřej Bojar$^{\ufalletter}$}
\affiliation{$^{\hspace{-0.5mm}\ethletter}$ETH Zürich, Department of Computer Science}
\affiliation{$^{\ufalletter}$Institute of Formal and Applied Linguistics, Faculty of Mathematics and Physics, Charles University\\
\text{}\hspace{1mm}{\{\hrefEmail{zouhar@ufal.mff.cuni.cz}{zouhar},\hrefEmail{kloudova@ufal.mff.cuni.cz}{kloudova},\hrefEmail{popel@ufal.mff.cuni.cz}{popel},\hrefEmail{bojar@ufal.mff.cuni.cz}{bojar}\}@ufal.mff.cuni.cz}
}
\end{authgrp}

\begin{abstract}
The overall translation quality reached by current machine translation (MT) systems for high-resourced language pairs is remarkably good.
Standard methods of evaluation are not suitable nor intended to uncover the many translation errors and quality deficiencies that still persist.
Furthermore, the quality of standard reference translations is commonly questioned and comparable quality levels have been reached by MT alone in several language pairs.
Navigating further research in these high-resource settings is thus difficult.
In this article, we propose a methodology for creating more reliable document-level human reference translations, called ``optimal reference translations,'' with the simple aim to raise the bar of what should be deemed ``human translation quality.''
We evaluate the obtained document-level optimal reference translations in comparison with ``standard'' ones, confirming a significant quality increase and also documenting the relationship between evaluation and translation editing.
\end{abstract}

\footnotetext{$^*$Apart from the affiliated institutions and the funding for this project (Ministry of Education, Youth and Sports of the Czech Republic LM2018101 LINDAT/CLARIAH-CZ), the last author is additionally funded by the 19-26934X grant of the Czech Science Foundation (NEUREM3) and the third author by the Horizon Europe Innovation grant no. 101070350 (HPLT).}

\maketitle

\vspace{-17mm}

\null\hfill
\raisebox{-0.1cm}{\includegraphics[width=0.35cm]{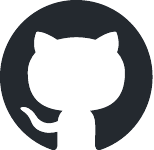}}
\hspace{-2mm}
\href{https://github.com/ufal/optimal-reference-translations}{\small \texttt{ github.com/ufal/optimal-reference-translations}}

\vspace{-3mm}
\null\hfill
\raisebox{-0.1cm}{\includegraphics[width=0.35cm]{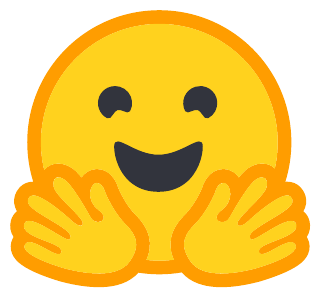}}
\href{https://huggingface.co/datasets/zouharvi/optimal-reference-translations}{\small \texttt{ hf.co/datasets/zouharvi/optimal-reference-translations}}

\vspace{0.3cm}

\section{Introduction}

Machine translation (MT) is routinely evaluated using various segment-level similarity metrics against one or more reference translations.
At the same time, reference translations acquired in the standard way are often criticized for their flaws of various types.
For several high-resourced language pairs, MT quality reaches levels comparable to the quality of the reference translation \citep{freitag2022results,hassan2018achieving} and sometimes MT even significantly surpasses humans in a particular evaluation setting \citep{popel2020transforming}.
Given this, one could conclude that state-of-the-art MT has reached the point where reference-based evaluation is no longer reliable and we have to resort to other methods (such as targeted expert evaluation of particular outputs), even if they are costly, subjective and possibly impossible to automate.

The narrow goal of the presented work is to allow for an ``extension of the expiry date'' for reference-based evaluation methods.
In a broader perspective, we want to formulate a methodology for creating reference translations which avoid the often-observed deficiencies of ``standard'' or ``professional'' reference translations, be it multiple interfering phenomena, inappropriate expressions, ignorance of topic-focus articulation (information structure) or other abundant shortcomings in the translation, indicating their authors' insensitivity to the topic itself, but above all to the source and target language.
To this end, we introduce so-called optimal reference translations (ORT), which are intended to represent optimal (ideal or excellent) human translations (should they be the subject of a translation quality evaluation).\footnote{While we have no formal proof that the translations actually reached this optimality (i.e. nothing can be better), we are confident that the result of translatologist collaboration comes close to this bar, especially given the evaluation results.}
We focus on document-level translation and evaluation, which is in line with current trends in MT research \citep{maruf2019survey,ma2020simple,gete2022tando,castilho2022dela} and also this special issue of NLE.
We hope that ORT will represent a new approach to the evaluation of excellent MT outputs by becoming a gold standard in the true sense of the word.
Our work is concerned with the following questions:
\begin{itemize}
\item How to navigate future MT research for languages for which the quality level of MT is already very good?
\item Is it worth creating an expensive optimal reference translation to compare with MT?
\item If various groups of annotators evaluate optimal reference and standard translations, will they all recognize the difference in quality?
\end{itemize}

\noindent
Subsequently, our contributions are:
\begin{itemize}
\item definition of optimal reference translation and an in-depth analysis of evaluations and the relationship between evaluation and translation editing;
\item reflection on what it means to be a high-quality translation for different types of annotators;
\item publication of the \emph{Optimal Reference Translations of English$\rightarrow$Czech} dataset with a subset evaluated in aforementioned manner.
\end{itemize}

After discussing related work in this context (\Cref{sec:related}), we focus on defining ORT and describe its creation process (\Cref{sec:ORT}).
Next, we describe our evaluation campaign of ORT, the data, annotation interface, and annotation instructions (\Cref{sec:annotation}).
We then turn to a statistical perspective of our data and measure the predictability of human ratings (e.g. \emph{Overall} rating from \emph{Spelling, Style, Meaning}, etc.) using automated metrics (\Cref{sec:statistical}).
We pay special attention to predicting document-level rating from segment-level.
In the penultimate \Cref{sec:qualitative_analysis}, we provide a detailed qualitative analysis of human annotations and discuss this work in the greater perspective of human evaluation of translations (\Cref{sec:discussion}).
Analysis code and collected data are publicly available.

\section{Related Work}
\label{sec:related}

Evaluating translations (machine or human) is without doubt an extremely demanding discipline. Researchers have recently contributed several possible ways to approach the evaluation of translation quality in high-resource settings. We focus on the latest findings in this area, which -- like our contribution -- look for a possible new direction where future translation quality evaluation can proceed. 
The presented study is primarily concerned with the evaluation of human translations (``standard'' vs. our optimal references) but the same evaluation methodology is applicable to machine translation.

Recently, \citet{freitag2022results} discussed metrics that were evaluated on how well they correlate with human ratings at the system and segment level.
They recommended using neural-based metrics instead of overlap metrics like BLEU which correlate poorly with human ratings, and demonstrated their
superiority across four different domains. 
Another relevant finding was that expert-based evaluation (MQM, Multidimensional Quality Metrics, \citet{lommel2014multidimensional}) is more reliable than DA (Direct Assessment, \citet{graham2013continuous}), as already confirmed by \citet{freitag2021experts}. The MQM method relies on a fine-grained error analysis and is used for quality assurance in the translation industry. 
\citet{popovic2020informative} proposed a novel method for manual evaluation of MT outputs based on marking issues in the translated text but not assigning any scores, nor classifying errors.
The advantage of this method is that it can be used in various settings (any genre/domain and language pair, any generated text).

Other unresolved issues in the field of translation evaluation include the question of whether it is better to evaluate in a source- or reference-based fashion.
As evidenced by e.g. \citet{kocmi2022findings}, reference-based human judgements are biased by unstable quality of references.
For some language pairs and directions, however, it is still the main method of assessment.
\citet{licht2022consistent} proposed a new scoring metric which is focused primarily on meaning and emphasises adequacy rather than fluency, for several reasons (e.g., meaning preservation is a pressing challenge for low resource language pairs and assessing fluency is much more subjective).

Methods for automatic human translation quality estimation exist \citep{specia2014predicting,yuan2018human}, though the field focuses primarily on machine translation quality estimation.
Furthermore, the definition of translation quality remains elusive and is plagued by subjectivity and low assessment agreement \citep{house2001translation,kunilovskaya2015far,guerberof2017quality}.

\section{Optimal Reference Translations}
\label{sec:ORT}

Our \pojem{optimal reference translation (ORT)} represents the ideal translation solution under the given conditions.
Its creation is accompanied by the following phases and factors:
\begin{itemize}
\item diversity at the beginning (multiple translations are available from different translators, i.e., in principle there are at least two independently-created translations available),
\item discussion among experienced translation theoreticians / linguists in search for the best possible solutions, leading to consensus,
\item editing the newly created translations, reaching a point where none of the translation creators comes up with a better solution.
\end{itemize}
Another important condition is the documentation of all stages of the translation creation (archiving the initial solutions, notes on shortcomings, suggestions for other potential solutions, notes on translation strategies and procedures, record of the discussion among the authors, reasons why a solution was rejected, record of the amount of time spent on each text, etc.).
The final characteristic of the creation of an optimal reference translation is the considerable amount of time spent by the creators on the analysis, discussion, and creation of new translations.
In our definition of ORT, optimality therefore refers to:
\begin{itemize}
\item a carefully thought-out and documented translation process, and
\item the quality of the resulting translation.
\end{itemize}

It however does not include the time aspect, in the sense of minimizing the time spent on the translation process.
This choice is likely one more key distinction from ``professional'' translation.
Incontestably, more than one version of ORT may be produced. %
The resulting ORT may vary depending on the individuality of its creators. %
Of course, the creators take into account the purpose and intended audience of ORT, just like in standard translations, but different collectives of ORT creators may perceive the intended purpose and audience differently or consider finer details of these aspects. Moreover, factors such as idiolect, age, experience, etc. can also play a large role, but unlike standard translations, there must always be a consensus among the creators of ORT.

\begin{figure}

\centering  
\resizebox{\linewidth}{!}{
\begin{tabular}{lp{15.5cm}}
\toprule
\textbf{Key}\hspace{-5mm} & \textbf{Text} \\
\midrule
SRC\hspace{-5mm} &
Professor Blair Grubb, Vice-Principal (Education) at the University, said: ``To get to this stage, our students and graduates faced competition from peers attending some of the world's top universities.'' \\
\cmidrule{2-2}
N1 &
Prorektor pro oblast vzdělávání profesor Blair Grubb prohlásil: „Aby se naši studenti dostali až sem, museli čelit konkurenci svých vrstevníků, kteří studují na nejlepších světových univerzitách.“ \\
& \textit{Vice-Principal for Education, professor Blair Grubb, said: ``To get to this point, our students have had to face competition from their peers studying at the world's best universities.''} \\
\cmidrule{2-2}
P1 & Profesor Blair Grubb, univerzitní zástupce ředitele pro vzdělávání, uvedl: „Abychom se dostali až do této fáze, museli naši studenti a absolventi čelit konkurenci svých vrstevníků, kteří studují na nejlepších světových univerzitách." \\
& \textit{Professor Blair Grubb, the University's Deputy Director of Education, said: ``For us, to get to this stage, our students and graduates have had to face competition from their peers studying at the world's best universities.''} \\
\cmidrule{2-2}
P2 &
Profesor Blair Grubb, zástupce děkana (vzdělávání) na univerzitě, řekl: ``Aby se dostali do této fáze, čelili naši studenti a absolventi konkurenci svým vrstevníkům, kteří navštěvují některé z nejlepších světových univerzit.'' \\
& \textit{Professor Blair Grubb, Associate Dean (Education) at the University, said: ``To get to this stage, our students and graduates have faced competition to their peers who attend some of the world's top universities.''} \\
\cmidrule{2-2}
P3 &
Profesor a zástupce ředitele pro vzdělávání Blair Brubb university uvedl: ``Aby se naši studenti a absolventi dostali do této fáze, museli čelit vrstevníkům z několika nejlepších universit světa.''\\
& \textit{Professor and Deputy Director of Education at Blair Brubb University said: ``To get to this stage, our students and graduates have had to face peers from several of the best universities in the world.''} \\
\bottomrule
\end{tabular}
}
\caption{Example translations of the same source into Czech. Literal transcriptions of the translations are shown in \textit{italics}. \textbf{N1}: translatologist collaboration (optimal translation), \textbf{P1}: professional translation agency (post-edited MT), \textbf{P2, P3}: professional translation agency.}
\label{fig:overview_sources}
\end{figure}

\subsection{Translation Creation}
\label{subsec:translation_creation}

The underlying dataset without the evaluation has already been described in Czech \citep{kloudova2022sas}. 
The 130 original English texts (news articles available from the Internet, covering topics ranging from politics and economics to sports and social events) were translated from English into Czech by three human translators for the Conference on Machine Translation 2020 (WMT20).
The three translators were hired by WMT organizers from a translation agency.
The resulting three independent parallel Czech translations (P1, P2, P3) serve as basic reference translations, from which a final ``optimal reference translation'' could be synthesized.
It was anticipated that our creators of ORT (two translators-cum-theoreticians -- professionals who deal with translation from both a practical and a theoretical point of view)\footnote{One of them is a co-author of this article. However, the translations are later independently evaluated and hence, to the best of our knowledge and conscience, we do not consider this to be a conflict of interest or otherwise a methodological flaw.} would always choose the best translation solutions from the existing three versions, or create new solutions if necessary. However, the available translations from the first-stage translators were often of insufficient quality. Therefore, in the creation of our optimal reference translations, more emphasis was placed on the input of the creators of the final version rather than on the synthesis of existing translations.

The process of creating our ORT can be described as follows: our tandem of translators-cum-theoreticians worked as a translator and revisor pair.
One of them produced a first version, which the other carefully compared with the original and critiqued if necessary. Notes on the first version of the translations were given in the form of comments on individual segments of the text.
The author of the first version of the translations subsequently accepted or, with justification (and subsequent discussion), did not accept the suggestions in the comments.
The crucial point in the discussion was always that the final solution should be fully in line with the beliefs of both translation authors.
It is worth mentioning that the discussion between the two creators had, to a large extent, the written form of exchanging notes.
ORT thus do not demand live, synchronous, attention of the creators.

The result of this process were two versions of ORT (many more versions could have evolved, though, our priority was not diversity, but above all quality – so we decided to create two version in parallel, N1 and N2).
The first version (denoted N1) is closer to the original both in terms of meaning and linguistic (especially syntactic) structure.
The second version (N2) is probably more readable, idiomatic and fluent, being even closer to the Czech news style, both syntactically, e.g. by emphasizing the ordering of syntactic elements typical of news reporting, and lexically, e.g. by a more varied choice of synonyms. The presented work is centered around the evaluation of the various human translations.
Because N2 has not been created for all segments of the translation (not all the original segments allowed an appropriate linguistic variation, i.e. N1 was identical to N2), we decided not to use it.
Thus, four translations were included in the evaluation – one optimal reference translation (N1) in addition to the three existing human translations.

In \Cref{fig:overview_sources}, we show the sources and example translations (P1, P2, P3), together with one of the two versions of our optimal translation, N1. During the evaluation, each translation can be further edited by annotators in which case we identify the resulting segment as e.g. as ``P1 EDIT by annotator A4.''
We will encounter examples in \Cref{sec:qualitative_analysis}.

\subsection{Annotation Campaign}
\label{sec:annotation}

\subsubsection{Annotators}

We hired 11 native Czech annotators for the evaluation of translations in three groups:
(1) four professional translators,\footnote{Defining who is a professional translator is not easy. The factors influencing the degree of professionalism of a translator include, among others, education and experience. Professional translators in our study have at least one of the following: (1) completed an M.A. degree programme in English-Czech translation studies, (2) completed an M.A. degree programme in interpreting or philology, or (3) have at least 10 years of translation experience.}
(2) four non-experts,
(3) three students of MA Study Programme Translation and/or Interpreting: Czech and English at the Institute for Translation Studies.\footnote{\href{https://utrl.ff.cuni.cz/en}{utrl.ff.cuni.cz}}
Their proficiency and end-campaign questionnaire responses are presented in \Cref{tab:questionnaire}.

\vspace{-3mm}
\subsubsection{Data}

Out of the original data (\Cref{subsec:translation_creation}), we randomly selected, with manual verification, 8 consecutive segments in 20 documents which were to be annotated.
We refer to these 8 segments as documents because they contain most of the documents' main points.
Each segment corresponds approximately to one sentence, though they are longer (31 source tokens on average) than what we would find typical for the news domain.
The data contain document-level phenomena (e.g. discourse), so segments can not be translated and evaluated independently.

\vspace{-3mm}
\subsubsection{Annotation Interface}
\begin{figure}[htbp]
\vspace{-3mm}
\includegraphics[width=\linewidth]{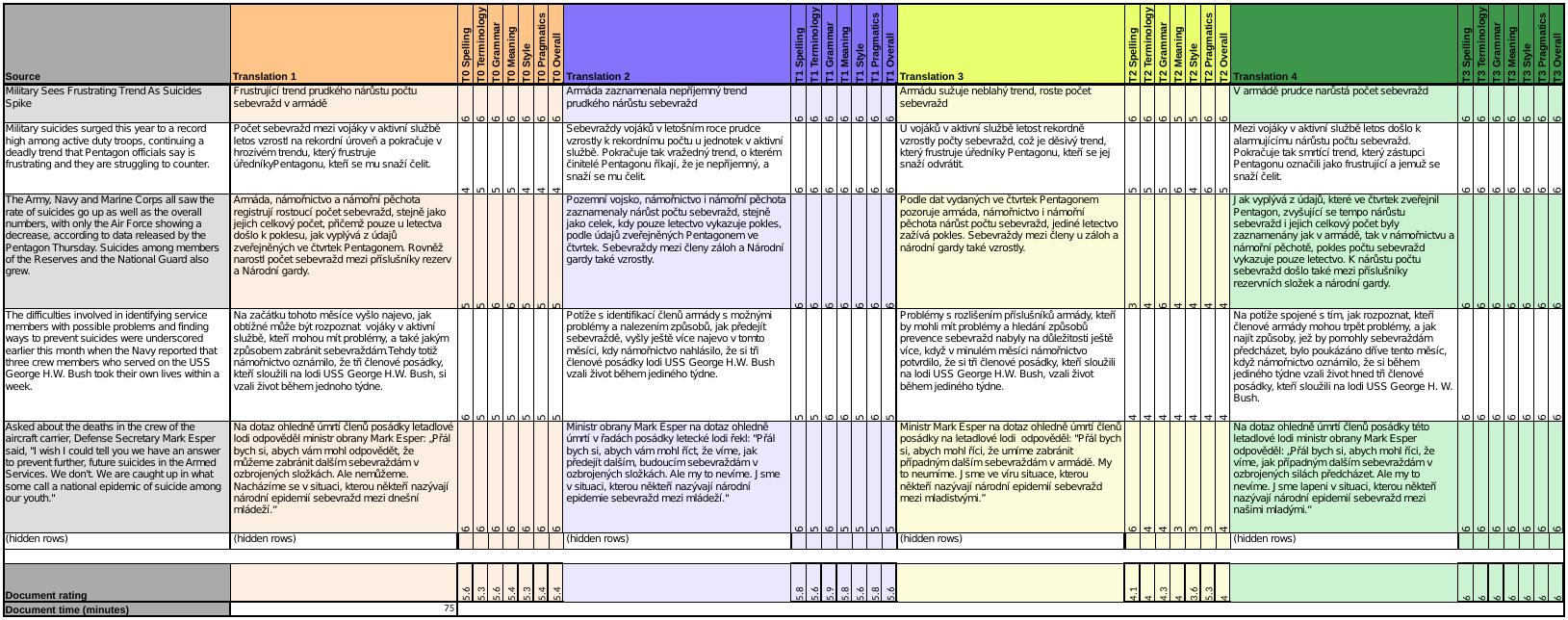}
\caption{First 5 rows of a screen for a single document with source and 4 translations in paralel. Screens were accessed by annotators in an online spreadsheet program. Note: Scalable graphics -- zoom in.}
\label{fig:annotation_ui}
\end{figure}

We provided the annotators with online spreadsheets which showed the source text and all four translation hypotheses.
This way each translation could be compared against the others while having the context available (e.g. to check for consistency).
Each hypothesis column was distinguished by a colour, as shown in \Cref{fig:annotation_ui}, and based on annotator feedback (\Cref{tab:questionnaire}), we believe that it was manageable to perform annotations despite the amount of information shown.
We showed the rest of segments in the source language for context but did not provide any translation hypotheses for the annotators to consult or rate.
Each of the 20 documents was shown in a separate tab/sheet.
The annotators worked on the evaluation in a span of 3 months in an uncontrolled environment.

\vspace{-3mm}
\subsubsection{Annotation Instructions}

The task for annotators was three-fold, see \Cref{sec:instructions_fulltext} for the full annotation guidelines.
\begin{itemize}
\item Grade each segment translation on a decimal scale from 0 (least) to 6 (most) in categories \emph{Spelling, Terminology, Grammar, Meaning, Style, Pragmatics} and \emph{Overall} (e.g. $4.0 or 5.8$). This scale was chosen to balance the number of attraction points for annotators (integers) and to contain a middle point (3).
\item Grade each document as a whole on the same scale and categories.
\item If a segment would not receive the highest grade, there would be something wrong in the translation. Therefore, the annotators should %
edit the hypothesis translation into a state to which they would give it the maximal scores.
\end{itemize}

\section{Quantitative Analysis}
\label{sec:statistical}

\subsection{Annotator Questionnaire}
\label{tab:questionnaire}

After the annotation campaign, the annotators filled a brief survey with questions about their perception of the task and their strategy.

We did not constrain the annotators in what order they should perform the annotations.
As a result, they employed various approaches, most popular being \emph{segment-category-translation}.\footnote{I.e. first finish all annotation categories in a translation, then all annotation categories in the second translation, etc., and afterwards move to the second segment.}
While we attempted to not introduce a bias, almost all annotators filled in categories one-by-one as they were organized in the user interface.\footnote{I.e. starting from \emph{Spelling} and ending with \emph{Overall}.}
This could have an effect on the rating.
For example, by establishing and drawing attention to the specific 6 features, the final \emph{Overall} rating may be influenced primarily by them and it would not have been if the ordering was reversed.
\emph{Pragmatics} and \emph{Overall} were reported as the hardest to evaluate, while \emph{Spelling} was the easiest, especially because errors in spelling can be seen even without deeper translatological analysis and there were not many of them in the translations.
The annotators self-reported utilizing the preceding and following context around half the time to check for document-level consistency.
While they proceeded mostly linearly, about 20\% (self-reported estimate) of previously completed segments were later changed.
We intentionally shuffled the ordering of translations (columns in each sheet) so that the annotators would not build a bias towards the translation source in e.g. the second column.
However, the annotators reported that despite this, they were sometimes able to recognize a specific translation source based on various artifacts, such as systematically not translating or localizing foreign names.

\subsection{Collected Annotations}
\label{total-collected-stats}

We do not do any preprocessing or filtering of the collected data.
This is justified by our all annotators working on the same set of documents and by the fact that we have established connections with each of the annotators and deem them trustworthy.
Any bias of an annotator's rating would therefore be present in all documents which would not hinder even absolute comparisons.
Nevertheless, we examine annotator variation later in this section.
In total for 20 documents, we collected:
\begin{itemize}
\item
7k segment-level annotations (1.8k annotations of 4 translation hypotheses).
Each hypothesis is %
edited unless it received a very high score (in 4k cases).
This amounts to 49k ratings across all categories.
\item
880 document-level annotations (220 annotations of 4 translation hypotheses.)
This amounts to 6.2k ratings across all categories.
\end{itemize}

\subsection{Quality of Initial Translations}

Recall the grading scale from 0 (least) to 6 (most).
The translation sources (P1, P2, and P3) were of varying quality, as shown in \Cref{fig:systems_seg_doc_level}.
Overwhelmingly, N1 was evaluated the highest followed by P1, P2 and P3, in this order.
Furthermore, there is a strong connection between the ratings on segment and document-level and also across evaluation categories.

\begin{figure}[htbp]
\centering
\includegraphics[width=\linewidth]{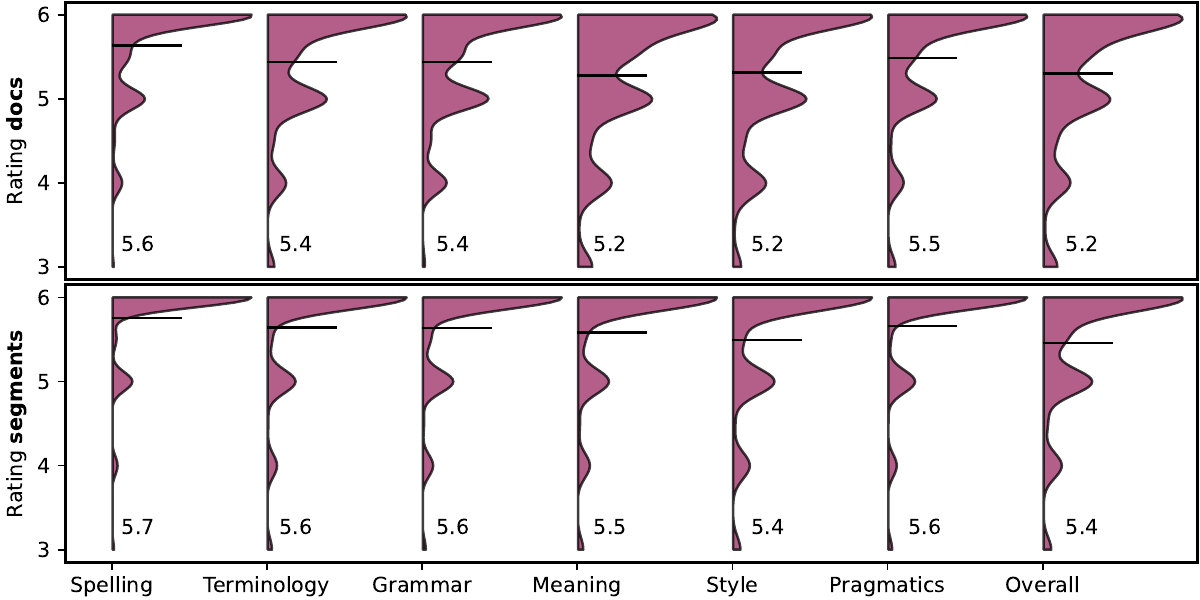}
\caption{Distribution densities of ratings of each collected variable (thin tail cropped $\geq$ 3 for higher resolution of high-density values). Numbers and horizontal lines show feature means.}
\label{fig:violin_features}
\end{figure}

The density distribution of features in \Cref{fig:violin_features} shows the natural tendency of annotators to use integer scores.
It also shows that all features are heavily skewed towards high scores and that on average documents receive lower scores than their segments.

\subsection{Inter-Annotator Agreement}
\label{sec:iaa}

To measure inter-annotator agreement, we aggregate pairwise annotator Pearson correlations on the segment-level.\footnote{Even though the data are not normally distributed, the Perason correlation reveals agreement controlled for each annotator's mean and variance.}
At first, this agreement is quite low ($\rho$ = 0.33).
It can however be explained upon closer inspection of agreement across translations.
While inter-annotator correlations for the the worst translation P3 were $\rho$ = 0.50, the best translation had $\rho$ = 0.13.
We hypothesize that with less variance and therefore signal for rating, the inter-annotator agreement drops.
This is even more visible from the pairwise annotator correlations for the \emph{Grammar} category, in which N1 has made almost no errors ($\rho$ = 0.03).
In 28\% of cases, the ordering of \emph{Overall} scores for segments was the same between pairs of annotators and in 66\% of cases they differed by only one transposition.
In other words, the difference in the score ordering was 2 positions or more only in 8\% of cases.
Further individual effects of annotators are discussed in \Cref{sec:annotator_differences}.

\subsection{Modelling Overall Quality from Components}

In this section we attempt to model the \emph{Overall} category based on individual categories, degree of %
translation editing and individual annotators.

\subsubsection{Other Categories Individually}
\label{sec:individual_categories}

\begin{figure*}[htbp]
\begin{minipage}{0.49\linewidth}
\begin{figure}[H]
\centering
\includegraphics[width=\linewidth]{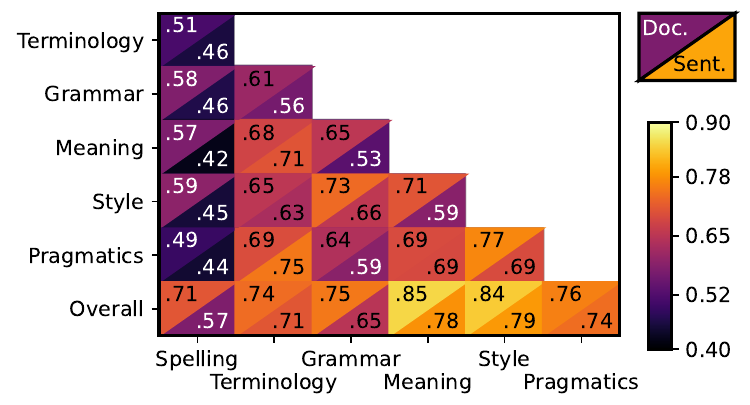}
\caption{Pearson correlations between individual features on document- (top-left) and segment- (bottom-right) level.}
\label{fig:category_correlation}
\end{figure}
\end{minipage}
\hfill
\begin{minipage}{0.49\linewidth}
\begin{figure}[H]
\centering
\includegraphics[width=\linewidth]{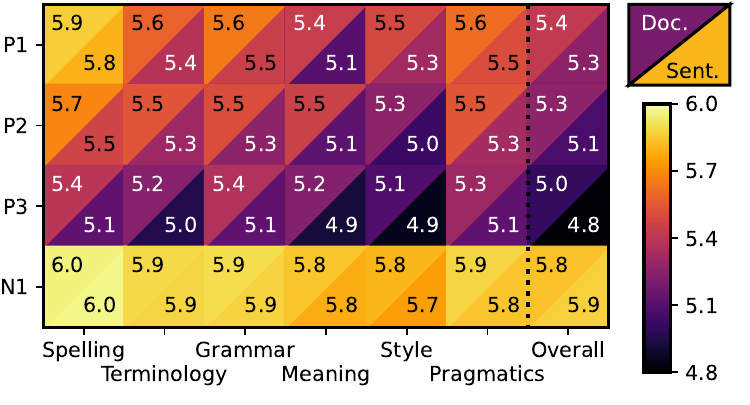}
\caption{Averages of ratings for different translation sources on document- (top-left) and segment- (bottom-right) level across features.}
\label{fig:systems_seg_doc_level}
\end{figure}
\end{minipage}
\end{figure*}

We first consider the predictability of individual categories and measure it using Pearson's correlation (0 = no relationship, 1 = perfect linear relationship).
For both the document and segment level, we observe similar correlations, see \Cref{fig:category_correlation}.
Notably spelling is much less predictive of other categories than the rest.
A possible explanation is that this was the least common mistake and the values are therefore concentrated around the highest possible score (\Cref{fig:violin_features}).
\emph{Overall} correlates the most with \emph{Meaning} and \emph{Style}.
This can be explained similarly because those features had the largest variances.

\subsubsection{Linear Regression on Other Categories}
\label{sec:linear_regression}

\begin{figure}[t]
\centering
\includegraphics[width=0.65\linewidth]{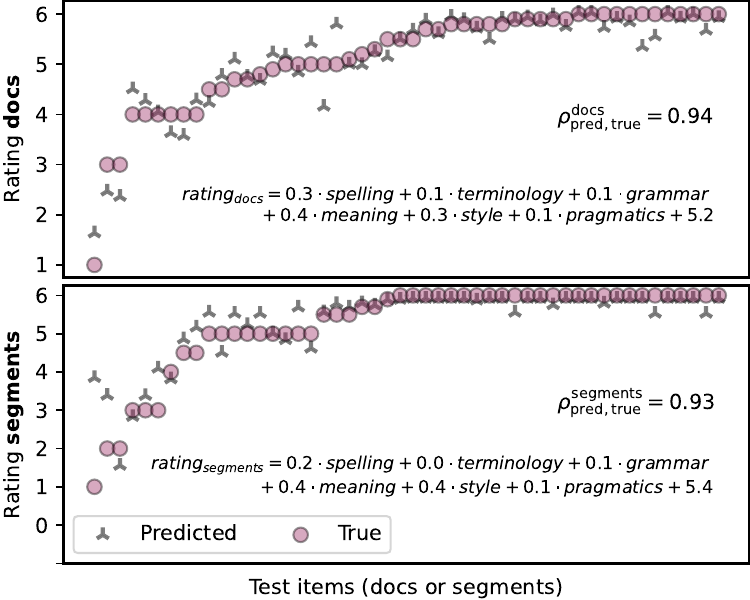}
\caption{Predictions of linear regression models (on document- and segment-level) for all test set items sorted by true \emph{Overall} score. Formulas show fitted coefficients and Pearson's correlations with the true scores.
Only a random subset of points shown for visibility.
}
\label{fig:lr_doc_seg}
\end{figure}

We treat the prediction of \emph{Overall} from other categories as a regression task with 6 numerical input features (\textit{Spelling}, \textit{Terminology}, etc) and one numerical output feature (\textit{Overall}).
We subtract the mean to preserve only the variance to be able to interpret the learned coefficients of a linear regression model.
We split document- and segment-level ratings into train/test as 778/100 and 6925/100, respectively.
\Cref{fig:lr_doc_seg} shows the results of fitting two linear regression models together with the coefficients of individual variables.
Because the distributions of features is similar, as documented in \Cref{fig:violin_features}, we can interpret the magnitude of the coefficient as the importance in determining the \emph{Overall} score.
For both the document and segment level, \emph{Spelling} and \emph{Meaning} have the highest impact while \emph{Terminology} and \emph{Style} have the least impact.%
\footnote{These interpretations are, however, not fully conclusive because of a possible latent co-dependent.
The \emph{Overall} variable may in reality be largely dependent on another variable $X$ for which we do not have annotations.
One hypothetical translation source could be very good if measured on the $X$ variable and also \emph{Overall} and unrelated to that also good in the \emph{Spelling} level, which would yield similar results to those presented.}
The linear regression model is further negatively affected by the non-linearity of the human bias towards round numbers, which the model is not able to take into consideration.
The fitted coefficients are confirmed by annotator reponses in the questionnaire in which
\emph{Meaning}, \emph{Style} and \emph{Pragmatics} was most important to them when evaluating \emph{Overall}. 

\begin{wrapfigure}{r}{0.55\linewidth}
\vspace{-15mm}
\centering
\includegraphics[width=\linewidth]{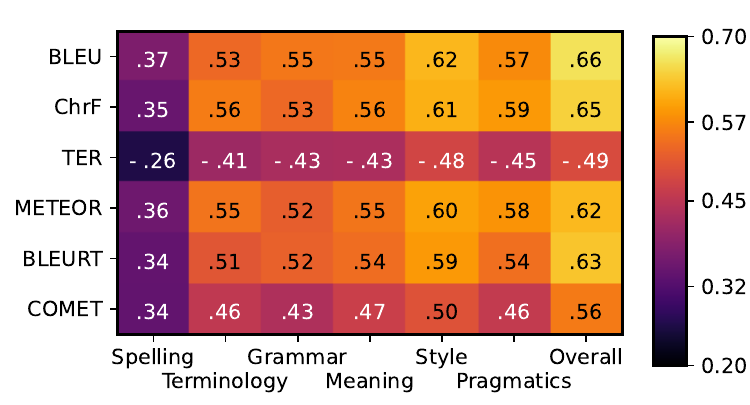}
\caption[Segment-level Pearson correlations between the collected scores and automated metrics between the original and %
edited versions of a segment. Color is based on absolute value of the correlation (note TER).]{Segment-level Pearson correlations between the collected scores and automated metrics between the original and %
edited versions of a segment. Color is based on absolute value of the correlation (note TER).\footnote{Most metrics are scored from e.g. 0 (lower quality) to 100 (higher quality). For TER it is the opposite (lower values mean higher quality). This explains the negative correlations.}
}
\label{fig:pe_score_correlation}
\vspace{-5mm}
\end{wrapfigure}

\subsubsection{Automated Metrics}
\label{sec:automated_metrics}

As mentioned in \Cref{sec:annotation}, annotators were tasked to post-edit texts to a state which they would be content with.
As a result, the annotators post-edited 62\% of all the segments on average.
We compute several automatic metric scores between the original and edited 
versions of segments and compare them to the collected scores, such as \emph{Overall}.
This allows us to answer the question: \emph{Does the post-edited distance (as measured by automated metrics) correspond to the annotator score (negatively)?}
The results in \Cref{fig:pe_score_correlation} show that there is very little difference between individual metrics.
Most score categories are equally predictive with the exception of \emph{Overall} (most) and \emph{Spelling} (least).
The explanation for this phenomena for \emph{Spelling} is again (\Cref{sec:iaa}) much lower variance.
Overall, the more the %
annotators changed the original text in their %
post-editing, the lower score they assigned to the hypothesis.
Including the metrics in the prediction of \emph{Overall} in \Cref{sec:linear_regression} does not provide any additional improvement on top of other categories (final segment-level $\rho$ is still $0.93$).

\subsection{Annotator Differences}
\label{sec:annotator_differences}

\begin{figure}[t]
\centering
\includegraphics[width=\linewidth]{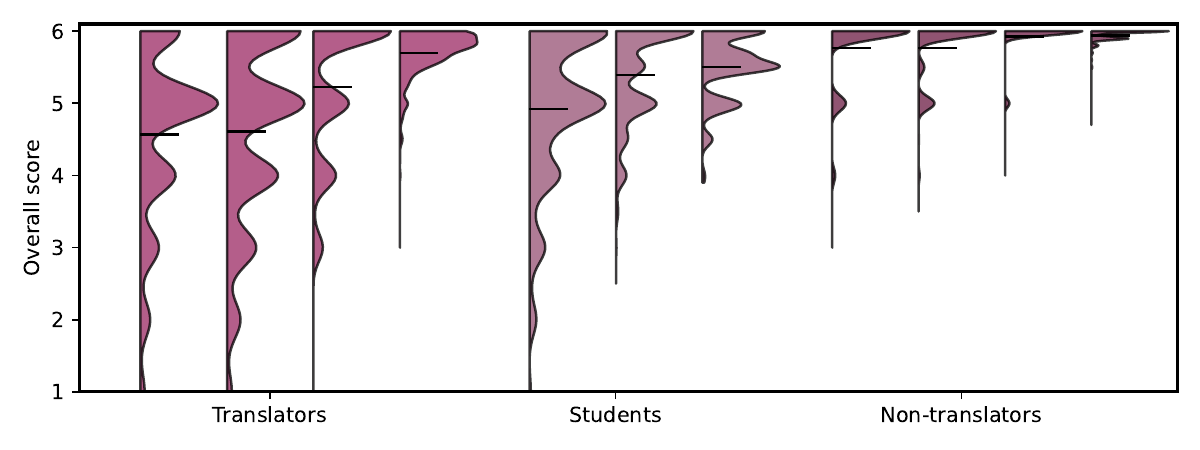}
\caption{Distribution densities of ratings of \emph{Overall} for individual annotators.}
\label{fig:annotator_dist}
\end{figure}

Recall that we considered three types of annotators: professional translators, students of translation and non-translators.
Despite the same annotation guidelines, their approach to the task was vastly different.
For example, \Cref{fig:annotator_dist} shows the distribution of segment-level ratings of \emph{Overall}.
Professional translators produced much more varying and spread-out distribution, especially compared to non-translators, who rated most segments very high.
The group differences should be taken into account when modelling the annotation process statistically.
When predicting segment-level \emph{Overall} from other categories, as in \Cref{sec:linear_regression}, the individual annotator Pearson correlation ranges from as high as $0.98$ to as low as $0.59$.
Similar to results of \citet{karpinska2021perils} we find that expert annotators are important and have less noise.
The average correlations with \emph{Overall} for the translator, student and non-translator groups are $0.93$, $0.91$ and $0.80$, respectively.
The expertise feature alone yields $0.36$ correlation with \emph{Overall} and users alone $0.45$.
This is expected as the groups and users have different means of the variable.
This information can be used in combination with other predictive features to push the segment-level correlation from $0.93$ (\Cref{fig:lr_doc_seg}) to $0.95$.
Greater improvement is achieved when combined with the %
editing distance, such as pushing BLEU from $0.66$ (\Cref{fig:pe_score_correlation}) to $0.76$ when individual annotators are considered as an input feature (one-hot encoded).

\subsection{Modelling Document-level Scores}

Our annotation instructions explicitly reminded annotators to always consider the context.
In other words, already our segment-level scores reflect the coherence and cohesion of the whole text, i.e, how the text is organized and structured in the previous and/or subsequent segments. This is a rather important difference from automatic segment-level evaluation which discards any context.
Annotators reported that in deciding document-level scores, they focused on the segments which were previously rated the lowest: that means, an individual poorly rated segment greatly influences the rating of the whole. We consider this observation essential for various future translation evaluations.
We confirm this with results in \Cref{tab:seg_to_doc} where the \emph{min} aggregation of segment-level ratings is a good prediction (comparable to or slightly better than \emph{avg}) of the document-level rating.
Based on segment-level ratings, we are able to predict document-level \emph{Overall} quality with $\rho = 0.71$.

It is worth noting that a similarly high correlation ($\rho = 0.70$) is achieved when predicting the document-level \emph{Style} from the corresponding segment-level ratings. This category was supposed to reflect also the coherence and cohesion of the document. Annotators saw the entire original text, but only evaluated certain translated segments. However, they were assumed to have read the entire source text and to use the information for their evaluation. This was reflected in the \emph{Style} category.

\textexample{
\label{txt:wimbledon}
\textbf{SOURCE}: \emph{The All England Club, which \examplehighlight{hosts the Wimbledon tournament}, handed the fine to \examplehighlight{Williams} after she reportedly caused damage during a practice round \examplehighlight{on the outside courts} on June 30, according to The Associated Press and CNN.}\\
\textbf{ORIG}: \emph{Klub All England Club, který \examplehighlight{hostí turnament}, udělil dle zpravodajů The Associated  Press a CNN \examplehighlight{Williams} pokutu poté, co 30. června údajně způsobila škodu během  cvičného kola \examplehighlight{venku na kurtech}.}\\
\textbf{EDITED}: \emph{Klub All England Club, který \examplehighlight{pořádá wimbledonský turnaj}, udělil dle zpravodajů The Associated Press a CNN \examplehighlight{Williamsové} pokutu poté, co 30. června údajně způsobila škodu během cvičného kola na \examplehighlight{venkovních kurtech}.}
}

In \Cref{txt:wimbledon}, the original translator (ORIG) did not consider the context of the whole document, translated only word for word and committed numerous interferences.
It is completely unusual in the context of Wimbledon to use the phrase \textit{``hostí turnament''} (hosts the tournament, both words being examples of lexical interference from English).
In the context of tennis, the phrase \textit{``venku na kurtech''} (out on the courts) is also unusual.
In Czech, feminine names names are typically marked using Czech morphology (e.g. \textit{Serena Williams $\rightarrow$ Serena Williamsová}), which is the form predominantly found in the press.
In this sentence, the name Williams follows the names The Associated Press and CNN, which is very confusing for the Czech reader.
The feminine form thus makes the whole text easier to interpret and understand.
The evaluator has correctly intervened in the text by using collocations such as \textit{``pořádá wimbledonský turnaj''} (hosts the Wimbledon tournament) and writes about \textit{``venkovní kurty''} (the outside courts) and uses the feminine form \textit{``Williamsová''}.
All these changes are highlighted in bold in \Cref{txt:wimbledon} and demonstrate the evaluator's (translation student) sense of textual continuity and their knowledge of the overall global context, which should have been the task of the original translator.

\begin{wrapfigure}{r}{0.55\linewidth}
\vspace{-8mm}
\centering
\resizebox{0.7\linewidth}{!}{
\begin{tabular}{lcccc}
\toprule
\textbf{Category} & \textbf{min} & \textbf{max} & \textbf{avg} & \textbf{med}\\
\midrule
{Spelling} &               0.38 &   0.17 &   0.36 &   0.30 \\
{Terminology} &            0.65 &   0.33 &   0.66 &   0.62 \\
{Grammar} &                0.64 &   0.36 &   0.66 &   0.60 \\
{Meaning} &                0.57 &   0.28 &   0.52 &   0.48 \\
{Style} &                  0.70 &   0.26 &   0.69 &   0.63 \\
{Pragmatics} &             0.66 &   0.42 &   0.69 &   0.64 \\
{Overall} &                0.71 &   0.32 &   0.68 &   0.61 \\
\bottomrule
\end{tabular}
}
\caption{Pearson correlations of predictions from segment-level aggregations to document-level scores. E.g. for the \emph{Overall} category with \emph{min} aggregation:
$\rho\big(\{d^\text{Overall}: d \in \mathcal{D}\}, \{\min\{s^\text{Overall}: s \in d\}: d \in \mathcal{D}\}\big)$.}
\label{tab:seg_to_doc}
\vspace{-10mm}
\end{wrapfigure}

\section{Qualitative Analysis}
\label{sec:qualitative_analysis}

If we take a closer look at the evaluation of all four translations by individual annotators, several types of qualitative comparisons can be made. We focus on the following two perspectives: characteristics of the segments (1) for which N1 scores worse than P\{1,2,3\} and (2) for which there are the biggest differences in ratings.
Even though we include example translations in Czech, we provide explanations in English which are self-contained and hence do not require any knowledge of Czech.

\subsection{N1 scores worse than P\{1,2,3\}.}

N1 was evaluated with the highest scores in comparison to P1, P2, and P3, across all assessed features (see \Cref{fig:systems_seg_doc_level}).
However, there is a small number of segments in N1 which were evaluated worse than those in P\{1,2,3\}.
For a better overview, the frequencies at which the translations P\{1,2,3\} were evaluated better than N1 in the \emph{Overall} category are: P1: 6.16\%, P2: 4.96\%, P3: 3.99\%. We selected these segments (for each category, not only for \emph{Overall}) and analysed them.
In most cases our analysis revealed that the evaluation and the related editing of the translation was conditioned by the erroneous judgment of the annotators, who did not check the correct wording/meaning/usage in Czech and were tempted by the source text and/or the wrong parallel translations P\{1,2,3\}.
In other words, the optimal reference translation N1 stood the test and our analysis confirmed its quality, rather than the evaluators' judgement.

Furthermore, we also encounter a reduced (imperfect) rating of some segments in N1, although no errors are apparent and no changes in the edited version have occurred in comparison to the original version.
This finding is valid for all evaluated categories without exception.
We list here a number of such segments with reduced (imperfect) rating for each category: \emph{Spelling}: 1.0\%, \emph{Terminology}: 1.8\%, \emph{Grammar}: 2.0\%, \emph{Meaning}: 1.7\%, \emph{Style}: 2.8\%, \emph{Pragmatics}: 1.5\%, \emph{Overall}: 0.5\%, any category: 5.4\%.
We perform a detailed qualitative analysis across all the seven rating categories.

\subsubsection{Spelling}

In the spelling category, the following segment in \Cref{txt:naredra} demonstrates an ignorance on the part of the annotator (non-translator) and failure to reflect the correct spelling and declension of the name \emph{Narendra Modi} in Czech (correctly in nominative singular: \emph{Naréndra Módí}) and of the Czech equivalent to the verb \emph{harass} (correctly: \emph{perzekvovat}, although it often appears incorrectly as \emph{perzekuovat} in the language usage). The proposed edits are wrong.

\textexample{
\label{txt:naredra}
\textbf{SOURCE}: \emph{Sources said the action was in line with Prime Minister \examplehighlight{Narendra Modi's} address to the nation [...] when he had said some black sheep in the tax administration may have misused their powers and \examplehighlight{harassed} taxpayers [...]}

\textbf{N1 ORIG} (rating: 4.0): \emph{Podle zdrojů akce souvisí s projevem premiéra \examplehighlight{Naréndry Módího} k národu [...] v němž prohlásil, že jisté černé ovce v systému daňové správy podle všeho zneužívaly své pravomoci a \examplehighlight{perzekvovaly} daňové poplatníky [...]}

\textbf{N1 EDIT}: \emph{Podle zdrojů akce souvisí s projevem premiéra \examplehighlight{Nara[!]ndra Modi} k národu [...] v němž prohlásil, že jisté černé ovce v systému daňové správy podle všeho zneužívaly své pravomoci a \examplehighlight{perzekuovaly} daňové poplatníky [...]}
}

There are more segments with incorrect or unnecessary spelling edits. Unfortunately, some annotators not only erroneously ``correct'' what is actually right, but also miscategorize the changes.
We find erroneously corrected morphology in this category, etc.
For example, for the source \emph{The man pleaded guilty to seven charges involving [...]} the correct structure \emph{Muž se přiznal k sedmi trestným činům {\normalfont (dative case)} týkajícím se {\normalfont (dative case)} [...]} has been edited to \emph{Muž se přiznal k sedmi trestným činům {\normalfont (dative case)} týkajících se {\normalfont (genitive case, grammatical incongruency)} [...].} 

It is quite surprising the extent to which annotators do not verify and follow up information, leaving errors in translations that are contained in the original. This is particularly evident in the spelling category. An example is a typo in the original: \emph{(Pete) Townsend} (correctly \emph{Townshend}; however, spelled correctly in the previous segment of the source text). P1 has \emph{Townsend}, P2 \emph{Townshed} [!], P3 \emph{Townsend}. N1 uses the corrected form \emph{Townshend}, but this form has been edited in the evaluation with the result \emph{Townsend}.

\clearpage

\begin{wrapfigure}{r}{0.45\linewidth}
\vspace{-10mm}
\textexample{
\label{txt:sony}
\textbf{SOURCE}: \emph{Sony, Disney \examplehighlight{Back To Work} On Third Spider-Man Film}

\textbf{N1 ORIG} (rating 3.0): \emph{Sony a Disney \examplehighlight{opět}$_{\text{(again)}}$ spolupracují na třetím filmu o Spider-Manovi}

\textbf{N1 EDIT}: \emph{Sony a Disney $\emptyset$ spolupracují na třetím filmu o Spider-Manovi}
}
\end{wrapfigure}

\subsubsection{Terminology}

In the terminology category we also detected unnecessary or erroneous corrections.
For example, the correction of the segment in \Cref{txt:sony} does not fall under terminology (and demonstrates, inter alia, the annotator's failure to verify the information; this time the annotator was actually a professional translator). The proposed edit is not correct/necessary.

\subsubsection{Grammar}

In the grammar category, the above mentioned segment (\emph{Sony, Disney Back To Work On Third Spider-Man Film}) plays an interesting role, rated also 3.0 in this category, without any other changes.
The proposed change (mentioned above) does not reflect grammar or spelling.
As it turned out, this segment achieved the same rating from this annotator in other categories, too, namely meaning, style, and overall quality.
It affects the meaning only, though, being an erroneous change.

We agree with some changes in syntax, e.g. in \Cref{txt:homes_flooded} (rating 5.0 for this segment by a student annotator).

\textexample{
\label{txt:homes_flooded}
\textbf{SOURCE}: \emph{\examplehighlight{Homes were flooded} and people waded through streets with water up to their knees in scenes normally seen only \examplehighlight{at the height of the monsoon}.}

\textbf{N1 ORIG}: \emph{\examplehighlight{Domy byly zaplavené} a na ulicích se lidé brodili po kolena vodou, což bývá běžně k vidění jen v době, \examplehighlight{kdy monzunové období vrcholí}.}

\textbf{N1 EDIT}: \emph{\examplehighlight{Byly zaplaveny některé domy} a na ulicích se lidé brodili po kolena ve vodě, což bývá běžně k vidění jen v době, \examplehighlight{kdy vrcholí monzunové období}.}
}

\subsubsection{Meaning}
In the meaning category, we observe several inconsistencies in evaluating translated segments for N1 vs P\{1,2,3\}.
For example, reduced rating for N1 (4.0, by a non-translator) occurs in the following segment in \Cref{txt:patient}, though, there are no changes in the edited version. Both the translations P1 and P2 score 6.0, even though there are several erroneous meaning units. The \emph{initial therapy} is \emph{počáteční léčba} in Czech, not \emph{vstupní}, and the verb \emph{require} does not mean here that the patient himself required the therapy but that his/her medical condition required it. The expression \emph{chief medical officer} refers to the Czech equivalent \emph{hlavní/vedoucí/vrchní lékař}, not \emph{ředitel resortu zdravotnictví} (= \emph{Director of the Ministry of Health}).

\textexample{
\label{txt:patient}
\textbf{SOURCE}: \emph{All but one patient had gone through \examplehighlight{initial therapy}. That patient \examplehighlight{did require} the recollection of stem cells, \examplehighlight{chief medical officer} James Stein said.}

\textbf{N1 ORIG} (rating 4.0): \emph{Všichni pacienti až na jednoho absolvovali \examplehighlight{počáteční léčbu}. U dotyčného pacienta \examplehighlight{bylo zapotřebí} provést nový odběr kmenových buněk, uvedl \examplehighlight{hlavní lékař} James Stein.}

\textbf{P1 ORIG} (rating 6.0): \emph{Kromě jednoho prošli všichni pacienti \examplehighlight{počáteční terapií}. Dotyčný pacient  \examplehighlight{požadoval} nový odběr kmenových buněk, uvedl \examplehighlight{hlavní lékař} James Stein.}

\textbf{P2 ORIG} (rating 6.0): \emph{Všichni kromě jednoho pacienta prošli \examplehighlight{vstupní léčbou}. Tento pacient \examplehighlight{vyžadoval} znovuodebrání kmenových buněk, uvedl \examplehighlight{ředitel resortu zdravotnictví} James Stein.}

\vspace{1mm}
\textbf{N1 EDIT} = N1 ORIG \quad
\textbf{P1 EDIT} = P1 ORIG \quad
\textbf{P2 EDIT} = P2 ORIG
}

\subsubsection{Style}
In the style category, we agree with some edits made for N1, for example in the following segment in \Cref{txt:using_ai} rated 4.0. However, the rating of P1 is 6.0, although there have been very similar modifications to the style in the edited version of P1 as those in N1, and, furthermore, the annotator (translator) uses the translation strategy proposed in N1 ORIG for his P1 EDIT version.

\textexample{
\label{txt:using_ai}
\textbf{SOURCE}: \emph{Using data and artificial intelligence to try and boost revenues is part of HSBC's broader push \examplehighlight{to squeeze more out of its large physical network and client data, a key priority for interim Chief Executive Noel Quinn}.}

\textbf{N1 ORIG}: \emph{Využití dat a umělé inteligence ke zvýšení příjmů je součástí \examplehighlight{širší strategie HSBC, která tak chce vytěžit více ze své rozsáhlé fyzické sítě a klientských dat, což je klíčovou prioritou prozatímního generálního ředitele Noela Quinna}.}

\textbf{N1 EDIT}: \emph{Využití dat a umělé inteligence ke zvýšení příjmů je součástí \examplehighlight{širší strategie HSBC, která tak chce ze své rozsáhlé fyzické sítě a klientských dat vytěžit víc. Jde o jednu z hlavních priorit prozatímního generálního ředitele Noela Quinna}.}

\textbf{P1 ORIG}: \emph{Využití dat a umělé inteligence ke zvýšení výnosů je součástí \examplehighlight{širšího tlaku na HSBC, aby vytěžila více ze své rozsáhlé fyzické sítě klientů a klientských dat, což je klíčovou prioritou dočasného generálního ředitele banky Noela Quinna}.}

\textbf{P1 EDIT}: \emph{Využití dat a umělé inteligence ke zvýšení výnosů je součástí \examplehighlight{širší strategie HSBC, která chce ze své rozsáhlé fyzické sítě klientů a z klientských dat vytěžit víc. Jde o hlavní prioritu dočasného generálního ředitele banky Noela Quinna}.}
}

\subsubsection{Pragmatics}
The evaluation in the category of pragmatics is also inconclusive and the analysis of N1 segments rated worse than P\{1,2,3\} segments does not provide any convincing data.
For example, one of the annotators (non-translator) evaluates \Cref{txt:thunberg} N1 5.0, whereas P3 is evaluated 6.0. However, the only change we find in the edited version of N1 is the elimination of the adjective \emph{nadšenou}, although \emph{nadšená chvála} is a typical collocation in Czech, in contrast to the rather unusual formulation (and too literal translation) \emph{zářná chvála} used in P3, which remained unchanged. Furthermore, \emph{New York Magazine} is usually used in the Czech media in its original, not translated form. Nevertheless, P3 uses \emph{New Yorský magazín} (the adjective does not even exist in Czech). Other inappropriate or non-existent word units used in P3 include: \emph{díky} (= \emph{thanks to}) used in a negative context, \emph{Gettysburgského projevu} (correctly: \emph{Gettysburského}, without \emph{g}), \emph{pro svůj historický význam} (correctly: \emph{pro jeho historický význam}).
The word order is not based on the principle of the Czech functional sentence perspective and is non-idiomatic and non-standard.
We could give many more similar examples.

\textexample{
\label{txt:thunberg}
\textbf{SOURCE}: \emph{Thunberg's grim pronouncements have earned her savage criticism, and \examplehighlight{glowing} praise. \examplehighlight{New York Magazine} called her "the Joan of Arc of climate change," while The Guardian ranked her speech alongside President Lincoln's \examplehighlight{Gettysburg Address} for \examplehighlight{its} historical significance.}

\textbf{N1 ORIG} (rating 5.0): \emph{Hrozivé výroky vynesly Thunbergové ostrou kritiku i \examplehighlight{nadšenou} chválu. \examplehighlight{New York Magazine} ji nazval „Johankou z Arku klimatických změn“, zatímco The Guardian zařadil její projev pro \examplehighlight{jeho} historický význam vedle \examplehighlight{projevu prezidenta Lincolna v Gettysburgu}.} 

\textbf{N1 EDIT}: \emph{Hrozivé výroky vynesly Thunbergové ostrou kritiku i $\emptyset$ chválu. \examplehighlight{New York Magazine} ji nazval „Johankou z Arku klimatických změn“, zatímco The Guardian zařadil její projev pro \examplehighlight{jeho} historický význam vedle \examplehighlight{projevu prezidenta Lincolna v Gettysburgu}.}

\textbf{P3 ORIG} (rating 6.0): \emph{Thunbergová si \examplehighlight{díky} svým hrozivým výrokům vysloužila divokou vlnu kritiky a \examplehighlight{zářnou} chválu. \examplehighlight{New Yorský magazín} ji nazval „Johankou z Arku klimatické změny,“ zatímco The Guardian zařadil její projev vedle \examplehighlight{Gettysburgského projevu} prezidenta Lincolna pro \examplehighlight{svůj} historický význam.} 

\textbf{P3 EDIT} = P3 ORIG
}

\subsubsection{Overall}

The overall category shows similar inconsistencies as described in previous aspects. The annotators often neglect formal, meaning and other errors, as shown above.
\Cref{txt:equality} shows that different types of errors in P2 and P3 have been ignored.
The annotator (non-translator) correctly substitutes the word \emph{kredit} for \emph{úvěr} in P3, but does not recognize the wrong structure \emph{pokud jde o dobré jméno} in P2: the Czech word \emph{kredit} (\emph{a sum of money (credit) or other value as a loan for a specified period of time for a specified consideration (e.g. interest})) can also have a colloquial meaning \emph{trust, respectability} which is not the case here. The collocation \emph{public accommodations} includes all services, i.e. not only accommodation, but also catering, cultural activities (public spaces and commercial services that are available to the general public, such as restaurants, theaters, and hotels). The Czech word \emph{služby} (= \emph{services}) is correct, not \emph{ubytování} (= \emph{accommodation}). The trickiest collocation of this segment is \emph{jury service} which is not \emph{soudnictví} (= \emph{judiciary}) in a general sense but, more specifically, \emph{účast v soudní porotě} (= \emph{participation in jury trials}). Leaving aside all the overlooked errors, the annotator evaluates P2 and P3 better than N1, although in N1 and P2 he/she made one change, in P3 two changes of a comparable nature.

\textexample{
\label{txt:equality}
\textbf{SOURCE}: \emph{The Equality Act would extend nondiscrimination protections to LGBTQ individuals \examplehighlight{in credit}, education, employment, housing, federal financial assistance, \examplehighlight{jury service} and \examplehighlight{public accommodations}.}

\textbf{N1 ORIG} (rating 4.0): \emph{Zákon o rovnosti by měl rozšířit ochranu proti diskriminaci také na příslušníky sexuálních a genderových menšin, a to v oblasti \examplehighlight{úvěrů,} vzdělávání, zaměstnání, bydlení, federální finanční pomoci, \examplehighlight{účasti v soudní porotě} a \examplehighlight{služeb}.} 

\textbf{N1 EDIT}: \emph{Zákon o rovnosti by měl rozšířit ochranu proti diskriminaci také na příslušníky sexuálních a genderových menšin, a to v oblasti \examplehighlight{úvěrů,} vzdělávání, zaměstnání, bydlení, federální finanční pomoci, \examplehighlight{soudnictví} a \examplehighlight{služeb}.}

\textbf{P2 ORIG} (rating 5.0): \emph{Zákon o rovnosti by měl rozšířit ochranu proti diskriminaci na LGBTQ jedince, \examplehighlight{pokud jde o dobré jméno}, vzdělání, zaměstnání, bydlení, federální finanční pomoc, \examplehighlight{činnost porotců} a \examplehighlight{veřejné ubytování}.} 

\textbf{P2 EDIT}: \emph{Zákon o rovnosti by měl rozšířit ochranu proti diskriminaci na LGBTQ jedince, \examplehighlight{pokud jde o dobré jméno}, vzdělání, zaměstnání, bydlení, federální finanční pomoc, \examplehighlight{soudnictví} a \examplehighlight{veřejné ubytování}.}

\textbf{P3 ORIG} (rating 5.0): \emph{Zákon o rovnosti by zajišťoval ochranu proti diskriminaci LGBTQ osobám \examplehighlight{v oblastech kreditu}, vzdělání, zaměstnání, bydlení, federální finanční asistence, výkonu poradce a \examplehighlight{veřejného ubytování}.} 

\textbf{P3 EDIT}: \emph{Zákon o rovnosti by zajišťoval ochranu proti diskriminaci LGBTQ osobám \examplehighlight{v oblastech úvěrů,} vzdělání, zaměstnání, bydlení, federální finanční asistence, \examplehighlight{výkonu soudnictví} a \examplehighlight{veřejného ubytování}.}
}

\begin{wrapfigure}{r}{0.5\linewidth}
\vspace{-10mm}
\newcommand{\aoneemph}[1]{\textbf{#1}}
\centering
\resizebox{0.7\linewidth}{!}{
\begin{tabular}{lcccc}
\toprule
\textbf{Category} & \aoneemph{\textbf{A1}} & \textbf{A2} & \textbf{A3} & \textbf{A4} \\
\midrule
Spelling & \aoneemph{0} & 6 & 6 & 6 \\
Terminology & \aoneemph{2} & 6 & 6 & 4 \\
Meaning & \aoneemph{3} & 5 & 6 & 6 \\
Pragmatics & \aoneemph{3} & 6 & 6 & 5 \\
Overall & \aoneemph{2} & 6 & 6 & 5 \\
\bottomrule
\end{tabular}
}
\vspace{-2mm}
\caption{Scores of subset of categories for a selected segment from the translation P3 by annotator A1 (translator) and annotators A\{2,3,4\}(non-translators).}

\label{tab:a1_a2_a3_a4_example}
\vspace{-5mm}
\end{wrapfigure}

\vspace{-10mm}
\subsection{Individual Differences in Ratings}

This perspective detects and analyzes segments with the biggest differences in ratings among annotators.
In all categories we find differences of at least 4.0 (Terminology and Grammar), 5.0 (Meaning, Style, Pragmatics, and Overall), or 6.0 (Spelling).

In this part, we would like to highlight selected segments with the biggest differences across relevant categories and focus on finding out the reasons for the observed disprepancies. 
The following segment in \Cref{txt:central} shows a very low rating and multiple changes in the edited version by annotator A1, whereas annotators A2, A3, and A4 evaluate it with (almost) best scores and overlook even obvious (to the authors and translatologists) mistakes.
In the example, we use subscripts to expressions of interests.

\textexample{
\label{txt:central}
\textbf{SOURCE}: \emph{\examplehighlight{The Central Board of Indirect Taxes and Customs (CBIC)} -- the \examplehighlight{agency} that oversees GST and \examplehighlight{import tax collections} -- compulsorily retired 15 senior officers under \examplehighlight{Fundamental Rule} 56 \examplehighlight{(J)} on corruption and other charges, official sources said.}

\textbf{P3 ORIG}: \emph{\examplehighlight{Ústřední komise nepřímých daní a cel (UKNDC$_{2,4}$)} — \examplehighlight{agentura$_3$,} která dohlíží na  daně ze zboží a služeb a \examplehighlight{vybrání vývozních dávek$_2$} — odvolala 15 vedoucích  úředníků na základě \examplehighlight{Základního Pravidla$_1$} 56 \examplehighlight{(J)$_{1,3}$} o korupci a jiných obviněních,  uvedly oficiální zdroje.}

\textbf{P3 EDIT by annotator A1} (translator): \emph{\examplehighlight{Ústřední rada pro nepřímé daně a cla (CBIC$_{2,4}$)} – \examplehighlight{instituce$_3$,} která dohlíží na výběr daně ze zboží a služeb a\examplehighlight{dovozních daní$_2$} – odvolala patnáct vysoce postavených úředníků na základě \examplehighlight{paragrafu$_1$} 56 \examplehighlight{písm. j)$_{1,3}$,} o korupci a jiných obviněních, uvedly oficiální zdroje.}

\textbf{P3 EDIT by annotator A2} (non-translator): \emph{\examplehighlight{Ústřední komise nepřímých daní a cel (CBIC$_{2,4}$)} — \examplehighlight{agentura$_3$,} která dohlíží na GST a \examplehighlight{daně z importu$_2$} — odvolala 15 vedoucích  úředníků na základě \examplehighlight{paragrafu$_1$} 56 \examplehighlight{(J)$_{1,3}$} o korupci a jiných obviněních,  uvedly oficiální zdroje.}

\textbf{P3 EDIT by annotator A3} (non-translator): \emph{\examplehighlight{Ústřední komise nepřímých daní a cel (UKNDC$_{2,4}$)} — \examplehighlight{agentura$_3$,} která dohlíží na  daně ze zboží a služeb a \examplehighlight{vybrání vývozních dávek$_2$} — odvolala 15 vedoucích  úředníků na základě \examplehighlight{Základního Pravidla$_1$} 56 \examplehighlight{(J)$_{1,3}$} o korupci a jiných obviněních,  uvedly oficiální zdroje.}

\textbf{P3 EDIT by annotator A4} (non-translator): \emph{\examplehighlight{Ústřední komise nepřímých daní a cel (CBIC$_{2,4}$)} — \examplehighlight{agentura$_3$,} která dohlíží na  daně ze zboží a služeb a \examplehighlight{vybrání vývozních dávek$_2$} — odvolala 15 vedoucích  úředníků na základě \examplehighlight{Základního Pravidla$_1$} 56 \examplehighlight{(j)$_{1,3}$} o korupci a jiných obviněních,  uvedly oficiální zdroje.}
}

Annotator A1 rightly notices errors in spelling (lower and upper case letters$_1$), terminology (name of the institution and other terms$_2$),  meaning (unclear and contradictory statements$_3$), pragmatics (dealing with foreign realia and abbreviations$_4$). 
These individual assessments are also reflected in the category \emph{Overall}.
On the other hand, the annotators A2, A3, A4 do not (mostly) notice the errors mentioned above, or just change the wording of the abbreviation or replace the translation with the original abbreviation (A2) while maintaining the best rating.
From our point of view, the correct annotator is A1 with their relevant, thoughtful and sensitive interventions in the text.

There are many segments with similarly unbalanced ratings in our evaluation.
As the analysis shows, the biggest problem is that some annotators fail to recognize most of the errors.
Problematic is also lowering the rating even though no changes were made in the edited version.
It is unclear whether the annotators simply did not pay enough attention to their task and whether they would have reached the same conclusions even after a more careful consideration of the whole task. 
Our qualitative analysis of selected segments confirms the findings presented in Figure 5: translators are the most rigorous and careful (average rating 5.00), students are slightly less attentive (5.3), and non-translators notice errors the least (5.8). 

\subsection{Document-level Phenomena}

In this section we present examples that show the extent to which authors of translations P1, P2, P3, and N1, and especially our evaluators have considered the context of the whole document.
We examined the evaluated documents, including the source text and all four translations, looking for evidence of apparent respect or disregard for the document-level context.

Documents in which certain terms occur that should be consistent throughout the text and/or should correspond in a meaningful way to the thematic and pragmatic area, appear to be appropriate material to demonstrate this (spans marked with \textit{1}).\footnote{We mark all related spans in the source as well as in all discussed translations even if they are correct, for easy comparison.}
These observations are in line with MT evaluation methods focused on terminology \citep{zouhar2020wmt20,semenov2022automated,agrawal2023findings}.
Another phenomenon to be observed might be a particular way of spelling words, which generally have two or more accepted spellings and the convention is just to achieve a consistent spelling throughout the document (spans marked with \textit{2}). 

Finally, for the topic-focus articulation, also called functional sentence perspective (spans marked with \textit{3}), it is crucial to respect the context of the whole document \citep{danes1974functional,sgall1986meaning,hajivcova2013topic}.
It is concerned with the distribution of information as determined by all meaningful elements, including context.
In \Cref{txt:china_conflict}, evaluated by a non-translator, we selectively document the distribution of the degrees of communicative dynamism over sentence elements in Czech, which determines the orientation or \textit{perspective} of the sentence \citep{firbas1992}.\footnote{The examples in \Cref{txt:uni_dundee} also contain other translation errors, such as incorrect name translation \textit{Grubb/Brubb} and not only those related to the document-level phenomena. Unless otherwise stated, the evaluator has left these errors in the translation even after editing. Since we present associated discussions in more detail within the previous sections, we do not elaborate on them at this point.}

Our first example in this section, \Cref{txt:therapy}, illustrates the disregard for proper terminology.
Translations P{1,2} and N1 were not edited. The evaluator was a non-translator.

\textexample{
\label{txt:therapy}
\textbf{SOURCE}: \textit{
All but one patient had gone through initial therapy. That patient \examplehighlight{did require} the recollection of stem cells, \examplehighlight{chief medical officer} James Stein said.
}

\textbf{P1 ORIG}: \textit{
Kromě jednoho prošli všichni pacienti počáteční terapií.
Dotyčný pacient \examplehighlight{požadoval}$_{1 \text{(insisted on)}}$ nový odběr kmenových buněk, uvedl \examplehighlight{hlavní lékař}$_{1 \text{\small(chief medical officer)}}$ James Stein.
}

\textbf{P2 ORIG}: \textit{
Všichni kromě jednoho pacienta prošli vstupní léčbou.
Tento pacient \examplehighlight{vyžadoval}$_{1 \text{(demanded)}}$ znovuodebrání kmenových buněk, uvedl \examplehighlight{ředitel resortu zdravotnictví}$_{1 \text{(Director of the Ministry of Health)}}$ James Stein.
}

\textbf{P3 ORIG}: \textit{
Každý, až na jedno pacienta, prošel počáteční terapií.
Tento pacient \examplehighlight{potřeboval}$_{1 \text{(needed)}}$ odebrání kmenových buněk, uvedl \examplehighlight{vrchní zdravotní důstojník}$_{1 \text{(chief medical officer in command)}}$ James Stein.
}

\textbf{P3 EDIT} (non-translator): \textit{
Každý, až na jedno pacienta, prošel počáteční terapií. Tento pacient \examplehighlight{vyžadoval}$_{1 \text{(demanded)}}$  odebrání kmenových buněk, uvedl \examplehighlight{hlavní lékař}$_{1, \text{(chief medical officer)}}$ James Stein.
}

\textbf{N1 ORIG}: \textit{
Všichni pacienti až na jednoho absolvovali počáteční léčbu. U dotyčného pacienta \examplehighlight{bylo zapotřebí}$_{1 \text{(it was necessary)}}$ provést nový odběr kmenových buněk, uvedl \examplehighlight{hlavní lékař}$_{1 \text{(chief medical officer)}}$ James Stein.
}
}

In the next example, evaluated by the same non-translator, P1, P2, and N1 are consistent in terminology and spelling in this document.

\textexample{
\label{txt:uni_dundee}
(individual evaluation segments are separated with \segmentseparator)

\textbf{SOURCE}: \textit{
Three \examplehighlight{University of Dundee} students have been named regional winners as top graduates in Europe. \segmentseparator
The \examplehighlight{University of Dundee} students were named as top graduates in their respective fields in Europe in the 2019 Global Undergraduate Awards, whilst five other students from the same \examplehighlight{university} were praised by the judges.
\segmentseparator
Professor Blair Grubb, \examplehighlight{Vice-Principal (Education) at the University}, said: ``To get to this stage, our students and graduates faced competition from peers attending some of the world's top \examplehighlight{universities}.''
\segmentseparator
``I would like to offer my warmest congratulations to Scott, Chester and Lola on this fantastic achievement, alongside the other \examplehighlight{Dundee} representatives who were highly commended.''
}

\textbf{P3 ORIG}: \textit{
Tři studenti \examplehighlight{Univerzity v Dundee}$_{1,2}$ byli jmenováni regionálními vítězi jakožto jedni z nejlepších absolventů v Evropě.
\segmentseparator
Studenti \examplehighlight{Dundeeské University}$_{1,2}$ byly jmenováni v soutěži Globální vysokoškolské ceny 2019 jakožto jedni z nejlepších absolventů ve svých příslušných oborech, zatímco porotci ocenili dalších pět studentů ze stejné \examplehighlight{university}$_{2}$.
\segmentseparator
Profesor a \examplehighlight{zástupce ředitele pro vzdělávání}$_{1}$ Blair Brubb university uvedl: „Aby se naši studenti a absolventi dostali do této fáze, museli čelit vrstevníkům z několika nejlepších \examplehighlight{universit}$_{2}$ světa.“
\segmentseparator
„Chtěl bych srdečně poblahopřát Scottovi, Chesterovi A Lole za jejich fantastické úspěchy a zároveň velmi pochválit i ostatní reprezentanty \examplehighlight{Dundee}$_{1}$.“
}

\textbf{P3 EDIT} (non-translator): \textit{
\ldots
\segmentseparator
\ldots
\segmentseparator
Profesor a \examplehighlight{zástupce ředitele pro vzdělávání}$_{1}$ Blair Brubb $\emptyset$ uvedl: „Aby se naši studenti a absolventi dostali do této fáze, museli čelit vrstevníkům z několika nejlepších \examplehighlight{universit}$_{2}$ světa.“
\segmentseparator
\ldots
}
}

Our third example in this section, \Cref{txt:china_conflict}, is represented by the article \textit{China Says It Didn't Fight Any War Nor Invaded Foreign Land} which discusses armed conflicts between China and other countries.

In Czech, it is common to put the adverbials of time at the beginning or in the middle of a sentence (depending on the meaning and function of other sentence elements).
When appearing at the end of a sentence, they become the focus of the statement, so the communicative dynamism and sentence continuity may get broken (\textit{in 1979 / v roce / roku 2017, in 1979 / v roce 1979}, spans \textit{3a}).
Based on the information in the previous text (\Cref{txt:china_conflict}), the diplomatic resolution (\textit{diplomatically resolved, vyřešen diplomaticky / diplomatickou cestou}) stands in contrast to the armed conflicts, so it represents the focus of the statement and should appear at the end of the sentence (after the verb) (spans \textit{3b}).
Finally, the states \textit{Vietnam, Malaysia, the Philippines, Brunei and Taiwan} should be placed at the end of the Czech sentence (this becomes evident after reading and understanding the entire document where we are introduced to information about which countries China has had conflicts with) (spans \textit{3c}).
The word \textit{claims} (\textit{vznášejí nároky, mají protinároky, mají opačné nároky, si činí nárok}) belongs to the topic of the statement.

\textexample{
\label{txt:china_conflict}
\textbf{Previous article content:}

\small 
China on Friday said it has not provoked a ``single war or conflict'' or ``invaded a single square'' of foreign land, skirting any reference to the 1962 war with India.
``China has always been dedicated to resolving territorial and maritime delimitation disputes through negotiation and consultation,'' stated an official white paper released, four days ahead of the country set to celebrate its 70th anniversary of the leadership of the ruling Communist Party of China (CPC) on October 1. 
``China safeguards world peace through real actions. Over the past 70 years, China has not provoked a single war or conflict, nor invaded a single square of foreign land,'' the paper titled ``China and the World in the New Era'' said.
The white paper, while highlighting the CPS's ``peaceful rise'' made no reference of the bloody 1962 war with India and the vast tracts of land, especially in the Aksai Chin area, occupied by China.
The Sino-India border dispute involving 3,488-km-long Line of Actual Control (LAC) remained unresolved.
China also claims Arunachal Pradesh as part of South Tibet, which India contests.
So far, the two countries held 21 rounds of Special Representatives talks to resolve the border dispute.

\textbf{SOURCE}: \textit{
Besides the 1962 war, India and China had a major military standoff at Doklam \examplehighlight{in 2017} when the People's Liberation Army (PLA) tried to lay a road close to India"s narrow Chicken Neck corridor connecting with the \examplehighlight{North-Eastern states} in an area also claimed by Bhutan.
\segmentseparator
It was finally \examplehighlight{diplomatically resolved} after which both sides pulled back their troops.
\segmentseparator
China also had a major military conflict with Vietnam \examplehighlight{in 1979}. China claims sovereignty over all of South China Sea. Vietnam, Malaysia, the Philippines, Brunei and Taiwan have counter claims.
}

\textbf{P1 ORIG}: \textit{
Kromě války v roce 1962 měly Indie a Čína velký konflifkt v Doklamu \examplehighlight{roku 2017}$_{3a}$, kdy se čínská lidová osvobozenská armáda (ČLOA) snažila postavit silnici blízko indického úzkého Kuřecího krku spojující \examplehighlight{severo-východní}$_1$ \examplehighlight{země}$_1$ na území, na které si dělá nárok i Bhutan.
\segmentseparator
Vše bylo zcela \examplehighlight{diplomaticky vyřešeno}$_{3b}$ poté, co obě strany stáhly svá vojska.
\segmentseparator
Čína měla také veliký vojenský konflikt s \examplehighlight{Vietnamem}$_{3c}$ \examplehighlight{v roce 1979}$_{3a}$.
Čína vyhlásila svrchovanost nad všemi moři od jihu Číny.
\examplehighlight{Vietnam, Malajsie, Filipíny, Brunei a Tchaj-wan}$_{3c}$ mají protinároky.
}

\textbf{P2 ORIG}: \textit{
Kromě války v roce 1962 hrozilo mezi Indií a Čínou vypuknutí většího vojenského konfliktu u Doklamské náhorní plošiny \examplehighlight{v roce 2017}$_{3a}$, když se Čínská lidová osvobozenecká armáda (PLA) pokusila vybudovat železnici poblíž úzkého indického koridoru Kuřecí krk, který spojuje \examplehighlight{západní a východní}$_1$ \examplehighlight{část Indie}$_1$ v oblasti nárokované také Bhútánem.
\segmentseparator
Nakonec bylo vše \examplehighlight{vyřešeno diplomatickou cestou}$_{3b}$ a obě strany stáhly své vojenské jednotky.
\segmentseparator
\examplehighlight{V roce 1979}$_{3a}$ došlo také k velkému vojenskému konfliktu mezi Čínou a \examplehighlight{Vietnamem}$_{3c}$.
Čína si nárokuje svrchovanost nad celým Jihočínským mořem.
\examplehighlight{Avšak Vietnam, Malajsie, Filipíny, Brunej a Tchaj-wan}$_{3c}$ také vznášejí územní nároky na tuto oblast.
}

\textbf{P3 ORIG}: \textit{
Kromě války v roce 1962 byla mezi Indií a Čínou velká vojenská patová situace v Doklamu \examplehighlight{v roce 2017}$_{3a}$, kdy se Lidová osvobozenecká armáda pokusila položit silnici v blízkosti úzkého indického koridoru Kuřecí krk, který Indii spojuje se \examplehighlight{severovýchodními}$_1$ \examplehighlight{státy}$_1$ v oblasti, kterou si také nárokuje Bhútán.
\segmentseparator
Konflikt byl nakonec \examplehighlight{diplomaticky vyřešen}$_{3b}$ a obě strany poté stáhly svá vojska.
\segmentseparator
Čína měla také větší vojenský konflikt \examplehighlight{s Vietnamem}$_{3c}$ \examplehighlight{v roce 1979}$_{3a}$.
Čína si nárokuje suverenitu nad celým Jihočínským mořem.
\examplehighlight{Vietnam, Malajsie, Filipíny, Brunej a Tchaj-wan}$_{3c}$ mají opačné nároky.
}

\textbf{N1 ORIG}: \textit{
Kromě války v roce 1962 hrozilo mezi Indií a Čínou vypuknutí většího ozbrojeného konfliktu \examplehighlight{v roce 2017}$_{3a}$ u Doklamské náhorní plošiny, když se Čínská lidová osvobozenecká armáda (ČLOA) pokusila postavit železnici poblíž úzkého indického koridoru Kuřecí krk, který spojuje \examplehighlight{severní a východní}$_1$ \examplehighlight{část Indie}$_1$ v oblasti nárokované také Bhútánem.
\segmentseparator
Konflikt byl nakonec \examplehighlight{vyřešen diplomatickou cestou}$_{3b}$, načež obě strany svá vojska stáhly.
\segmentseparator
\examplehighlight{V roce 1979}$_{3a}$ došlo k většímu vojenskému konfliktu také mezi Čínou a \examplehighlight{Vietnamem}$_{3c}$. Čína si nárokuje svrchovanost nad celým Jihočínským mořem, ovšem na tuto oblast si činí nárok také \examplehighlight{Vietnam, Malajsie, Filipíny, Brunej a Tchaj-wan}$_{3c}$.
}
}

\section{Discussion}
\label{sec:discussion}

Evaluating optimal reference translation(s) is in many ways a more difficult task than evaluating a ``standard'' (human or machine) translation.
It is already a common practice in the translation industry to have multiple workers included in a translation of a single document (e.g. initial translator and quality assurance translator).
Based on our analysis of the optimal reference translation evaluation, it turns out that it is very crucial who evaluates such translations: Do the annotators have professional translation experience, or are they students of translation, or laymen in the field?
It appears that laypeople are less able to notice even critical mistakes in translations.
As a result for quality assurance, hiring only annotators with lots of translating experience seems to be a requirement.

However, important to determine is who the translation is for.
If it is for a wide audience who do not scrutinize the translation quality, it may not be worth the extra cost to hire highly skilled translation evaluators.
In turn, for evaluation of machine translation systems that have reached this very high level of quality, highly skilled evaluators are needed.

We do note, however, that perfect translations or annotations likely do not exist, only their approximations.
The cost of uncovering more translation errors is likely hyperlinear -- i.e. two rounds of annotations do not uncover twice as many mistakes.
Each use-case should therefore make explicit what the target quality level is and adjust the annotation protocol accordingly.

\section{Conclusion}

We defined the concept of optimal reference translation (ORT), geared towards regaining informative results in reference-based machine translation evaluation.
We then performed a careful manual evaluation and post-editing of ORT in comparison with three standard professional translation.
The evaluation confirms that ORT deserve their name and can be regarded as a truly golden reference.
In fact, the few times when ORT did not score best were examples of errors in this follow-up annotation, not examples of ORT deficiencies.
Additionally, we documented that manual evaluation at these high levels of quality can \textbf{not} be delegated to inexperienced annotators.
Only people with substantial translation experience are sensitive to the subtle differences and can provide qualified judgements.

\vspace{-2mm}
\subsection*{Future work}
While we focused on evaluating human translations, the idential setup could be used for evaluating MT models, which we plan to address in future work \citep{zouhar2024quality}.
This is not part of the present work which is focused on showing that the reference translations usually used are of insufficient quality and need to be reconsidered.
Our next step will be to assess which of the multitude of automatic metrics of MT quality are sensitive to the subtleties captured in our ORT and can thus be used to reliably evaluate MT outputs of high quality.
This will again require careful expert manual evaluation.

\clearpage

\bibliography{misc/bibliography.bib}
\bibliographystyle{misc/nlelike}
\label{lastpage}

\clearpage

\section{Annotation Guidelines}
\label{sec:instructions_fulltext}
\newcommand{\parheaderstyle}[1]{\vspace{-5mm}\subsubsection*{#1}}

The following is the main part of instructions which were distributed to the annotators.

{\small
\parheaderstyle{Introduction}

The goal of this study is to annotate the translation quality in seven categories. There are 20 documents in the shared Google sheet, marked as Edit1, Edit2 etc. (Orig1, … are described later in the text). The first column contains the source text in English, followed by four Czech translations. However, only eight segments should be evaluated in each document. If you don’t see a translation for some segments, it is not meant to be evaluated.
You will evaluate the translations both at the segment level and at the level of whole documents (or at the level of the eight continuous segments). You will also indicate a better translation if you are not satisfied with the current version.
Please read the source text first. The following is a possible evaluation procedure, but it is up to you how you proceed. The next steps are (for individual translations): 1. reading the translation, 2. evaluating the segments, 3. evaluating the whole document, 4. editing the segments so that you are satisfied with the translations, 5. reading the entire newly created text and possibly making minor changes. Please keep in mind that although you are also evaluating the segments separately, they are always part of a larger text, so you should pay special attention to how they relate to each other, i.e., also to the coherence and cohesion of the whole text. This should also be reflected in the assessment (category ``style'' below).

\parheaderstyle{Evaluation of segments}

Rate each of the four translations in the following seven categories on a scale from 0 (worst) to 6 (best):
\begin{itemize}[noitemsep]
\item spelling, punctuation, typography, typos,
\item terminology (correctness, consistency, normativity),
\item grammar: morphology (word forms) and syntax (sentence structure, functional sentence perspective),
\item meaning accuracy (mistranslation, addition, omission, untranslated text segment etc.),
\item style (appropriateness, consistency, idiomaticity, cross-sentence coherence and cohesion),
\item pragmatics (culture-specific reference, locale conventions, appropriateness for the Czech reader),
\item overall quality (evaluation of the translation in all the above-mentioned categories).
\end{itemize}

\parheaderstyle{Important notes}

You can rate from 0 (the worst rating) to 6 (the best rating); in addition to whole numbers (0, 1, 2, 3, 4, 5, 6), decimal numbers with one decimal place (e.g. 0.1 or 4.5) are allowed. It is not necessarily the goal to use the full range of ratings for individual translations, i.e., if you do not see an error in a given category (even if the translation of the rated segment is very easy and does not pose a challenge for the translator), you will rate the highest possible score (6). We leave it to the discretion of each evaluator to decide how serious they consider a particular error to be and how many points to deduct for it. If an error affects more than one category (typically, e.g., both categories 3 and 4), this should result in a reduced rating in all relevant categories.

\parheaderstyle{Evaluation of documents}

Rate the entire translation at the document level in the seven categories (the same as above for segment evaluation) on a scale from 0 to 6 (the same conditions as above for segment evaluation). The rating of the whole document is on the last line of each sheet.

\parheaderstyle{Editing of translated segments}

If a segment translation does not receive the highest rating (6) in overall quality, please edit the translation with minimal editing (changes, corrections) to the state that you would give the highest rating (6). To clarify, if translations 1, 2, 3, and 4 get an overall quality rating of 6, 5, 3, 6, respectively (for particular segments), you must edit translations 2 and 3 independently. The resulting translations should be based on the original translations, i.e., most of the time they will be different from each other even after your edits. You can use dictionaries or search the internet, but please do not use any machine translation systems. If possible, try not to copy text segments from previous translations, even if you like them.
Since you probably weren't satisfied with some of the translations and didn't give them the highest possible rating, you have edited some segments. For comparison, you can look at the original translation (OrigT), which is in another sheet. For example, for document 3, the sheet is called Orig3 and is listed just after Edit3. Edit only the EditT sheet.

\parheaderstyle{Time range}

To process one document in all four translations takes on average 25–75 minutes. Please indicate the time spent on the annotation of each document (in minutes) in the appropriate box in each sheet. If you are systematically outside this range, send us an email. Please note that annotating the first document usually takes much more time than annotating subsequent documents.

}

\end{document}